\documentclass[pmlr,twocolumn,10pt]{jmlr} 

\usepackage{listings}

\usepackage{booktabs}
\usepackage{float}
\usepackage{siunitx}

\usepackage[switch]{lineno}

\usepackage{xcolor}
\definecolor{tred}{RGB}{227, 11, 92}

\theorembodyfont{\upshape}
\theoremheaderfont{\scshape}
\theorempostheader{:}
\theoremsep{\newline}

\jmlrvolume{333}
\jmlryear{2026}
\jmlrsubmitted{LEAVE UNSET}
\jmlrpublished{LEAVE UNSET}
\jmlrworkshop{Conference on Health, Inference, and Learning (CHIL) 2026} 

\title[Can Language Models Identify Side Effects of Breast Cancer Radiation Treatments?]{Can Language Models Identify Side Effects of Breast Cancer Radiation Treatments?}

\author{%
\Name{Natalie Seah} \Email{cva3zw@virginia.edu}\\
\addr University of Virginia
\AND
\Name{Danielle S. Bitterman} \Email{dbitterman@bwh.harvard.edu}\\
\addr Mass General Brigham, Dana-Farber Cancer Institute, Harvard Medical School
\AND
\Name{Daphna Spiegel} \Email{dyspiegel@bwh.harvard.edu}\\
\addr Mass General Brigham, Dana-Farber Cancer Institute, Harvard Medical School
\AND
\Name{Thomas Hartvigsen} \Email{hartvigsen@virginia.edu}\\
\addr University of Virginia
}

\begin{document}

\maketitle

\begin{abstract}
Accurately communicating the side effects of cancer treatments to cancer survivors is critical, particularly in settings such as informed consent, where clinicians must clearly and comprehensively convey potential treatment toxicities. However, this task remains challenging due to clinical knowledge deficits about adverse treatment effects and fragmentation across electronic health record (EHR) systems. Large language models (LLMs) have the potential to assist in this task, though their reliability in oncology survivorship contexts remains poorly understood. We present a deployment-oriented stress-testing framework for evaluating LLM-generated radiation side effect lists in breast cancer treatment and survivorship care. Using 21 breast cancer patient profiles, we construct paired patient clinical scenarios that differ only in radiotherapy regimens to evaluate seven instruction-tuned LLMs under multiple prompting regimes. We then compare LLM outputs to a clinician-curated reference derived from informed consent documents at two major academic medical centers and developed by a team including more than seven breast radiation oncologists. The reference maps radiation dose-fractionation, fields, and locations to associated toxicities, broken down by frequency and temporal onset. Across models, we reveal sensitivity to minor documentation changes, trade-offs between precision and recall, and systematic under-recall of rare and long-term side effects. When used alone, constraints on the number of side effects generated reduce precision, and grounding outputs in clinician-curated side effect lists substantially improves reliability and robustness. These findings highlight important limitations of LLM use in oncology and suggest practical design choices for safer and more informative survivorship-focused applications.
\end{abstract}

\paragraph*{Data and Code Availability}
This study utilizes breast cancer patient profiles generated from two publicly available sources: (1) synthetic cancer patient scenarios from Chen et al. \citep{chen2023impact}, and (2) synthetic breast cancer scenarios developed by breast cancer specialists \citep{palepu2025exploring}. The exact profiles used in our experiments, along with the ground-truth side-effect file for model evaluation, are provided in the supplementary materials. All code for generating model responses and evaluating outputs can be found at \url{https://github.com/hartvigsen-group/llm-radiation-side-effects-eval}.

\paragraph*{Institutional Review Board (IRB)}
This research does not require IRB approval.

\section{Introduction}
Advances in cancer screening and treatment have led to a rapidly growing population of long-term cancer survivors, emphasizing the importance of effective post-treatment care \citep{Tonorezos2024Prevalence}. Survivorship care encompasses the ongoing monitoring, surveillance, and management of the long-term physical, psychological, and social effects of cancer and its treatments, making side-effect surveillance and management an essential component \citep{Mollica2025Defining}. In the United States, an estimated 18.6 million people were living with a history of cancer as of early 2025 \citep{wagle2025cancer}, a population expected to exceed 26 million by 2040 \citep{Tonorezos2024Prevalence}. As survivors transition from oncology specialists to primary care, accurate communication of prior treatments and their potential short- and long-term side-effects becomes increasingly important for specific clinical tasks such as drafting informed consent documentation and survivorship care plans \citep{grunfeld2010interface, ke2024decision}.
However, fragmented survivorship-relevant information across clinical documentation systems alongside clinical knowledge deficits about the long-term effects of cancer and its treatments make it difficult to successfully relay up-to-date tailored treatment effect information to cancer survivors \citep{alfano2022engaging, nekhlyudov2017integrating, nathan2013family}.

Large language models (LLMs) can potentially summarize and communicate cancer treatment effects to cancer survivors \citep{bitterman2024promise}.
However, their reliability in oncology remains in question \citep{chen2025medical}. For example, while LLMs can generate fluent and seemingly-relevant text, prior work has shown that they are prone to hallucinations \citep{yoon2025navigating}, factual inconsistencies \citep{singhal2023large}, and sensitivity to minor changes in input phrasing \citep{bitterman2024promise} in oncology contexts. As a result, understanding how LLM outputs vary when patient documentation differs in the level of treatment detail provided (e.g., general mentions of radiation therapy versus specific type or location), across different prompting strategies, and along clinically meaningful dimensions such as side-effect frequency and temporal onset is critical for assessing their safety and utility in oncology and survivorship care.

Cancer treatments often cause clinically meaningful side effects, which can be severe or chronic and can limit therapy tolerance and long-term survival \citep{gegechkori2017long, harrington2017late}. Despite this, the oncology workforce is insufficient to meet the needs of this expanding survivorship population \citep{alfano2022engaging}, and primary care providers frequently report inadequate knowledge of side effect surveillance and management \citep{vos2024primary}. Patients themselves also report receiving limited information about survivorship issues, including the potential risks of side effects \citep{ross2022still}. Together, these factors make it challenging to ensure that patients and providers have the information needed to manage side effects effectively and deliver high-quality survivorship care.

Given the broad range of cancers, treatments, and individual patient factors, it is not feasible to evaluate every possible scenario. To make the problem tractable while still addressing a clinically important context, we focus on breast cancer radiation therapy. Breast cancer is the most commonly diagnosed cancer in U.S. women and among the most prevalent cancers globally \citep{sung2021global,siegel2026cancer}. Radiation therapy is a cornerstone of breast cancer treatment \citep{boyages2018evolution}, and its side effects are important drivers of long-term survival and quality of life. It is also highly specialized. Side-effect risks vary with radiation dose, anatomical location, and technique, and this knowledge is not widely known even among physicians in other specialties \citep{wang2021radiation, siau2021non}. 

This highlights the need for improved ways to communicate patient-specific risks for precision decision-support and patient-tailored education. Currently, such tools for educating both patients and clinicians about treatment-related side effects are largely lacking \citep{kivistik2025perceptions}. Large language models (LLMs) have the potential to fill this gap, as they can support the generation of patient-facing informed consent documentation, where clinicians must clearly communicate the full spectrum of potential treatment toxicities, including rare and long-term effects, in a concise and structured manner \citep{Shi2025Transforming}. More broadly, such capabilities may also support survivorship education and clinical decision-making \citep{bitterman2024promise}. Given the growing interest in applying LLMs to health information tasks, rigorous methods are needed to evaluate their reliability and clinical utility in this context.

In practice, oncologists drafting informed consent documentation or survivorship care plans must accurately enumerate the potential toxicities of a patient's specific treatment. This is a task where missing a rare or long-term effect carries real clinical consequences. To evaluate whether LLMs can reliably support this workflow, we propose a stress-testing framework to measure how accurately LLMs identify side effects of breast cancer radiation treatments.
This framework is straightforward to apply to other cancers and treatments given sufficient clinician input.
We curate 21 breast cancer patient profiles that include their treatments and demographic information. 
We then create sets of perturbed profiles for each, adding an anatomical location to their treatment while keeping all other clinical information fixed.
Practicing oncologists then define evidence-based ground truth side effects, mapping breast cancer radiation types and anatomical locations to their established side effects. 
Each side effect is further annotated according to how commonly they are observed and also whether they are known to be short-term or long-term effects. 
This structure lets us evaluate survivorship-relevant distinctions, such as rare or delayed toxicities, alongside measures of side effect accuracy.
Using this stress-testing framework, we assess LLM robustness to input changes under multiple prompting regimes including free-form generation, list size constraints (i.e., limiting the number of side effects generated), and selection from a clinician-curated side-effect list.

Our experiments on seven state-of-the-art LLMs surface several consistent patterns in model behavior. First, the trade-off between precision and recall varies significantly across models: some models produce broader lists with moderate recall but low precision, while others achieve high precision at the cost of missing many clinically relevant side effects. 
Second, constraining the size of the generated side-effect list consistently reduces precision and has only minor, inconsistent effects on recall. 
Third, LLMs tend to under-recall rare and long-term radiation side effects, even when common short-term effects are reliably captured. 
Fourth, minor documentation changes, such as specifying radiation location, lead to substantial shifts in generated side-effect lists. 
Overall, grounding outputs in a clinician-curated side-effect list substantially improves precision, recall, and overall reliability across all models.

Our work makes four contributions. The primary contribution is the introduction of a stress-testing framework for evaluating LLM reliability in oncology applications that emphasizes clinically meaningful perturbations and survivorship-relevant distinctions rather than abstract benchmarks. The framework is designed to be reusable across models, and can be adapted to other cancer types given clinically grounded relationships between treatments and side effects.
Second, we present the first structured, experimental evaluation of how prompting strategies, documentation specificity, and side-effect characteristics together influence LLM performance in breast cancer radiation side-effect generation. 
Third, we demonstrate that grounding LLM outputs in expert-curated side-effect vocabularies substantially improves reliability, offering a practical design recommendation for safer use of LLMs in survivorship-related information tasks. 
Fourth, we produce a publicly-available dataset containing patient profiles paired with an oncologist-annotated set of breast cancer radiation treatments and their best-known side effects.
These resources can help foster further work in this important area, ultimately facilitating robust and reliable use of LLMs for cancer survivorship informational needs.

\section{Related Work}
\subsection{LLMs for Oncology}
Reviews of AI and large language models in oncology highlight multiple potential applications. For patients, LLMs have been proposed to support remote symptom monitoring, provide psychosocial support, simplify clinical language for easier understanding, and help navigate care by summarizing recommendations and follow-up plans \citep{chen2025large, bitterman2024promise}. Recent work has also explored multiple ways LLMs can improve patient comprehension and education: GPT‑4 can generate plain-language translations of clinical notes, helping patients understand their disease course and management \citep{kumar2025cross}, and other LLMs have been used to produce informed consent forms that maintain accuracy while improving readability, understandability, and actionability \citep{Shi2025Transforming}. 

In clinician-facing settings, LLMs have been explored for clinical decision support, summarizing literature, and extracting symptoms and adverse events from clinical notes \citep{chen2025large, bitterman2024promise, ferber2025development}. For example, prior work has shown that LLM-generated treatment recommendations can outperform very junior trainees but still lag behind experienced oncologists, illustrating both the potential and current limitations of LLMs for supporting clinical decision-making \citep{palepu2025exploring}. Beyond treatment recommendations, LLMs have also been applied to classify treatment-related toxicities and assist with clinical documentation workflows, suggesting they may streamline aspects of clinical care and improve access to relevant information, while underscoring the need for careful validation before deployment \citep{chen2025large, ruiz2024leveraging}.

\subsection{Evaluation of LLMs in Oncology Contexts}
Evaluation of LLMs in oncology primarily emphasizes comparison against established clinical knowledge or curated questions. Studies that test LLMs using a fixed set of medical exam questions show that current models can answer oncology-related questions with moderate to high accuracy, with the best models getting a large portion of the questions correct \citep{longwell2024performance}. However, analyses of errors reveal clinically meaningful inaccuracies even when overall accuracy appears high, underscoring the need for caution in interpreting performance metrics.

Existing reviews of LLMs in oncology show that evaluation approaches differ widely across studies. Studies span a diverse set of tasks, including summarizing trial results, generating patient educational materials, and predicting treatment side effects, and rely on varied data sources such as synthetic patient profiles, clinical notes, and structured EHR data \citep{chen2025large, mehan2025development}. Evaluation metrics range from standard quantitative measures (e.g., accuracy, precision/recall) to customized, task-specific scoring systems and qualitative assessments \citep{mehan2025development, carl2024large}. A comprehensive systematic review of LLMs in cancer decision-making across 15 cancer types found that most studies prioritize response accuracy and appropriateness, while comparatively few assess safety, potential harm, or clarity—dimensions essential for clinical reliability \citep{hao2025large}.
These findings indicate that, although LLMs can perform well on curated tasks, more standardized and clinically realistic evaluation methods are needed to assess their behavior in real-world oncology applications \citep{chen2025large, mehan2025development}.

Despite the diversity of evaluation approaches, few studies examine patient-facing survivorship tasks, such as side-effect informational content generation, or systematically test model robustness to variations in input, including documentation specificity or prompt phrasing. In response, our work introduces a stress-testing framework that evaluates the accuracy of LLM-generated side effects for breast cancer radiation treatments under controlled input perturbations.

\section{Methods}
We design a stress-testing framework to evaluate the reliability of large language models (LLMs) for oncology side-effect generation, using breast cancer radiation therapy as a representative high-stakes clinical domain. Our evaluation focuses on clinically-realistic variations in documentation specificity and prompting strategies that are commonly encountered in real-world use. 
Specifically, we assess how LLM-generated radiation side-effect lists vary as a function of (1) input documentation specificity, (2) output constraints imposed by prompting strategies, and (3) clinical characteristics of side effects, including frequency of occurrence and temporal onset. Our evaluation is grounded in a clinician-authored reference that maps breast radiation types and anatomical locations to known side effects, enabling systematic measurement of hallucination risk, recall gaps, and sensitivity to clinically meaningful perturbations.

\subsection{Patient Profiles} 
We construct a dataset containing patient profiles, each of which is a structured text summary that includes eight fields stored in an EHR-styled format.
In total, we curate 21 breast cancer patient profiles that include a history of radiation therapy by extracting the most-relevant profiles from two publicly available sources:
First, we extract the 20 AIME synthetic patient profiles \citep{palepu2025exploring} that explicitly mention radiation therapy. We only exclude two profiles, one of which radiation therapy had not yet begun, and one involving radiation for brain metastases, which would require substantially different side-effect mappings. This results in 18 breast cancer radiation profiles from AIME. Second, we include the three breast cancer profiles from OncQA \citep{chen2023impact}, totalling 21 patient profiles.

To ensure consistency across sources, we normalize all patient profiles into a shared EHR-style format used by OncQA. Each profile therefore contains eight fields: Age, Gender, Cancer Diagnoses, Past Medical History, Prior Cancer Treatments, Current Medication List, and Summary of Most Recent Oncology Visit.
Each field has corresponding text that is specific to each patient.

Some AIME profiles lack completion status of prior treatments, timing of the most recent oncology visit, and/or current medications.
Therefore, we populate missing fields by instructing GPT-5 to add clinically-plausible information that is consistent with the rest of the patient record. 
This augmentation was used only to ensure each patient has some relevant text for each field, and was validated to have never introduced new cancer diagnoses or radiation details.

\subsection{Radiation Specification Perturbation}
To evaluate sensitivity to documentation specificity, we create paired versions of each patient profile:

\begin{itemize}
    \item \textbf{Base profile:} Mentions radiation therapy without specifying radiation type or anatomical target.
    \item \textbf{Specified profile:} Adds a radiation type or breast anatomical location after a mention of ``radiation'' (e.g., ``radiation'' $\rightarrow$ ``radiation (left breast, chest wall)''). Otherwise identical to the base profile
\end{itemize}

In both base and specified profiles, radiation is described as part of the patient’s treatment course without introducing additional treatment events. The specified radiation type or anatomical location is appended directly to the existing mention of “radiation” (e.g., “radiation” → “radiation (chest wall and nodes)”) and is interpreted as referring to the same course of radiation therapy described in the profile. Thus, the prompts ask the model to identify side effects of the patient’s radiation treatment as described, rather than introducing a new or subsequent course of radiation. No profiles involve reirradiation scenarios.

Radiation types and locations are drawn from a clinician-authored reference file created by a practicing oncologist, which enumerates breast radiation modalities and anatomical targets along with their associated side effects. Specifically, our reference includes the following five types/locations:

\begin{itemize}
    \item Accelerated Partial Breast Irradiation (APBI)
    \item Chest Wall
    \item Chest Wall and Nodes
    \item Breast and Nodes
    \item Breast Only
\end{itemize}

Applying these specifications to the 21 base patient profiles results in 21 paired specified profiles (one per patient). No other clinical information is modified between paired profiles, allowing isolation of the effect of documentation specificity on LLM outputs.

\subsection{Ground Truth Side-Effect Reference}
Ground truth side effects are defined using a clinician-curated oncology reference derived from informed consent forms at two major academic medical centers, developed by a multidisciplinary team including more than seven breast radiation oncologists. These side effects reflect established risks of breast radiation supported by clinical literature rather than subjective opinion \cite{citrin2026effects, jagsi2015differences, shaitelman2015acute}.
Each side effect is additionally annotated along two clinical dimensions. 
First, each side effect is annotated as one of four \textit{frequency} categories: Common, Uncommon, Rare, or Extremely Rare.
Second, each side effect is annotated as one of two \textit{temporal onset} categories: Short-term or Long-term.
This structure enables evaluation not only of overall correctness but also of survivorship-relevant distinctions, such as whether models preferentially recall common acute effects while missing rare or delayed toxicities.

\subsection{Prompting Regimes (Stress Tests)}
We design four prompting regimes to reflect common deployment and prompt-engineering practices:

\begin{enumerate}
    \item \textbf{Free-form generation:} Models are instructed to list relevant radiation side effects based on the patient profile, without additional constraints. Prompts refer to “the patient’s radiation treatment,” requiring the model to infer side effects from the radiation described within the profile rather than introducing a separate treatment scenario.
    \item \textbf{Free-form generation with list-size constraint:} Models are instructed to list 20--30 side effects. This range was chosen based on the clinician-curated reference, which indicates that most breast radiation profiles are associated with approximately 20--30 side effects.
    \item \textbf{Selection from predefined list:} Models are provided the full clinician-curated side-effect vocabulary and asked to select those that are relevant to the given patient profile. Although the full set of possible side effects is visible, the model must still perform contextual filtering to identify which effects apply to a specific clinical scenario.
    \item \textbf{Selection with list-size constraint:} Models are provided the predefined side-effect list and instructed to select 20--30 relevant side effects, reflecting the typical number of side effects per profile in the reference.
\end{enumerate}

All prompts instruct the model to output only a bulleted list of side effects, with no explanatory text. The 20--30 side-effect range was determined from the ground truth reference and represents the typical number of side effects per breast radiation profile, ensuring that list-size constraints reflect realistic clinical expectations.
Our exact prompts used can be found in Appendix A. 

\subsection{Output Normalization and Matching}
To evaluate correctness, we use an LLM-based semantic matching procedure. For each model output, a separate LLM is prompted to identify which predicted side effects match entries in the ground truth list, allowing for differences in wording but requiring equivalence in clinical meaning. The matching model is instructed to output only the list of matched side effects, minimizing ambiguity. To validate this approach, we manually evaluated LLM match decisions on 30 examples by flagging false positives and false negatives, and found 96\% agreement with human judgments. This suggests that LLM-based matching is highly reliable for this task and does not materially affect our results.

\subsection{Evaluation Metrics}
For each patient profile and LLM output, let $\hat{H}$ denote the set of side effects generated by the model and let $S$ denote the ground truth side-effect set from the clinician-curated reference.
Using these ground truth labels, we measure standard Precision, Recall, and F1 scores.
Precision measures the proportion of generated side effects that are correct, recall measures the proportion of ground truth side effects captured, and F1 is the harmonic mean of precision and recall, providing a balanced assessment of accuracy. 

Precision, Recall, and F1 scores require ground truths. 
But new side effects are frequently identified, and some are debated.
So we also propose measuring an Overlap Ratio to examine robustness in model predictions to minor documentation changes. We first let $\hat{H}_\text{base}$ and $\hat{H}_\text{spec}$ be the side-effect sets generated from the base and specified profiles, respectively. Then, we can compute
    \[
    \text{Overlap ratio} = \frac{|\hat{H}_\text{base} \cap \hat{H}_\text{spec}|}{|\hat{H}_\text{base} \cup \hat{H}_\text{spec}|}
    \]
The overlap ratio thus quantifies how consistent a model’s outputs are when changes are introduced in the patient profile, reflecting the model’s robustness to variations in clinical documentation that could affect reliability.
Intuitively, a high overlap ratio indicates consistent outputs despite documentation changes; a low value signals sensitivity to minor profile perturbations.

    
    
    

After computing each metric for all patient profiles, models, and prompting regimes, we also compare models according to each side-effect's frequency and temporal onset to assess survivorship-relevant performance disparities.

\begin{figure*}[hbt!]
    \centering
    \includegraphics[width=\textwidth]{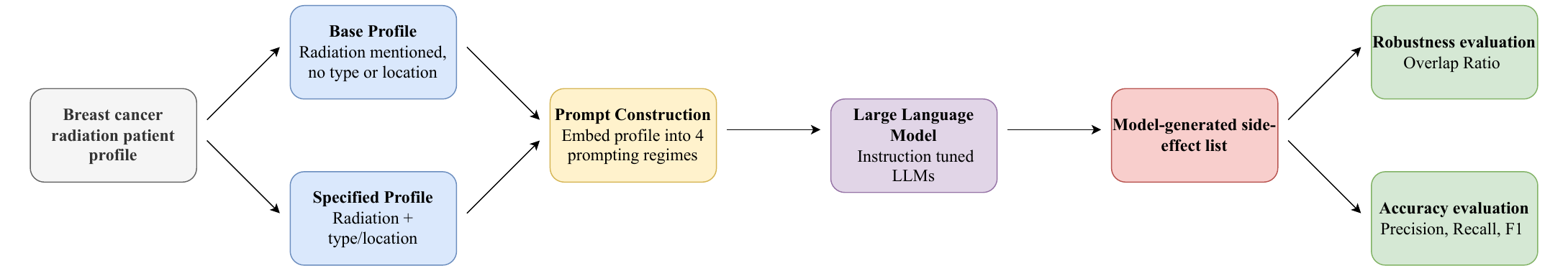}
    \caption{Deployment-oriented evaluation framework. Breast cancer patient profiles are constructed in paired base and specified forms that differ only in radiation documentation specificity. Profiles are converted into prompts and passed to large language models, which generate side-effect lists. Outputs are evaluated along two axes: robustness to documentation perturbations and accuracy relative to a clinician-curated reference.}
    \label{fig:flowchart}
\end{figure*}

\section{Experiments}
\subsection{Language Models and Inference}
We evaluate seven popular instruction-tuned LLMs that represent a range of architectures and training regimes: LLaMA-3.1-8B-Instruct, Gemma-2-9B-IT, Mistral-7B-Instruct-v0.3, Qwen-2.5-7B-Instruct, Phi-3-Medium-4K-Instruct, O4-Mini, and GPT-5.3. 
All are accessed from HuggingFace except for o4-mini and GPT-5.3, for which we use OpenAI's API.
We select these models as popular representatives of common instruction-tuned LLMs that are frequently deployed in research and applied settings, rather than to compare architectural innovations or maximize absolute performance.
All models were evaluated using identical prompts and a maximum of 1000 tokens. Specifically, HuggingFace models were generated with \texttt{do\_sample=False} (greedy decoding) to ensure deterministic outputs. For o4-mini and GPT-5.3, accessed via the Azure OpenAI API, sampling parameters such as temperature and top\_p are not supported; these models rely on fixed internal decoding behavior as determined by OpenAI.

Models were prompted using standardized zero-shot templates and generated a single response per prompt. Prompts explicitly instructed models to output only a bulleted list of side effects, with no explanatory text or additional commentary. All bullet characters were mapped to ``•'' for consistency across outputs.

\subsection{Implementation Details}
To isolate sensitivity to documentation specificity, patient profiles are constructed in paired form, differing only in whether radiation therapy was described generically or augmented with a specific radiation type or anatomical location. Radiation specifications are inserted programmatically via string replacement from a fixed set of clinician-approved breast radiation modalities and targets.

Model outputs are post-processed to normalize bullet formatting and casing prior to evaluation. Side-effect correctness is assessed using LLM-assisted semantic matching: For each model output, a separate instance of O4-Mini was prompted to identify which predicted side effects matched entries in a clinician-curated reference list, allowing for differences in wording while requiring equivalence in clinical meaning. The evaluation model was instructed to return only matching side-effect names, and matches were further restricted to the set of model-predicted items to prevent spurious or inflated alignments.

Precision, recall, and F1 scores are then computed based on these matched sets relative to the clinician reference for the corresponding radiation type. Overlap ratios were computed between outputs generated from base and specified profiles using set-based intersection over union.

\subsection{Free-Form Side-Effect Generation}

We first evaluate model behavior under unconstrained free-form generation, reflecting likely real-world usage when LLMs are asked to summarize radiation side effects without structured guidance.

\begin{table*}[t]
\floatconts
  {tab:sideeffect-metrics}

  {\small\begin{tabular}{lccc}
  \toprule
  \bfseries Model & \bfseries Precision & \bfseries Recall & \bfseries F1 \\
  \midrule
  LLaMA-3.1-8B   & 0.15 (0.13, 0.16) & 0.47 (0.42, 0.53) & 0.22 (0.19, 0.25) \\
  Gemma-2-9B     & 0.81 (0.76, 0.87) & 0.25 (0.23, 0.26) & 0.38 (0.36, 0.40) \\
  Mistral-7B     & 0.63 (0.57, 0.69) & 0.38 (0.35, 0.41) & 0.47 (0.44, 0.51) \\
  Qwen-2.5-7B    & 0.37 (0.31, 0.43) & 0.37 (0.31, 0.44) & 0.34 (0.30, 0.39) \\
  Phi-3-Medium   & 0.64 (0.52, 0.75) & 0.37 (0.33, 0.43) & 0.42 (0.38, 0.45) \\
  O4-Mini        & 0.79 (0.74, 0.84) & 0.57 (0.55, 0.60) & 0.66 (0.63, 0.69) \\
  GPT-5.3        & 0.79 (0.75, 0.83) & 0.60 (0.58, 0.62) & 0.68 (0.66, 0.69) \\
  
  \bottomrule
  \end{tabular}}  
  {\caption{Precision, Recall, and F1 Scores for Free-Form Side-Effect Generation. Values reported as mean (95\% bootstrapped CI) across 21 patient profiles.}}
\end{table*}

As shown in Table 1, model behavior under free-form generation varies substantially. Some models, such as Gemma-2-9B, O4-Mini, and GPT-5.3 achieve relatively high precision (0.79--0.81) but exhibit lower recall, indicating that they correctly generate side effects but miss many that are clinically relevant. Conversely, models such as LLaMA-3.1-8B produce broader lists with moderate recall (0.47) but very low precision (0.15), suggesting that many generated side effects are hallucinated or not applicable to the specified radiation type or anatomical target. For example, several models generated anatomically irrelevant toxicities such as pelvic radiation-associated cystitis and hematuria when evaluating breast radiation profiles, indicating hallucination of clinically implausible side effects.

Overall, free-form generation exposes a trade-off between precision and recall: some models favor accuracy at the cost of coverage, while others favor coverage at the cost of including unsupported side effects. These findings highlight the potential reliability risks of unconstrained LLM outputs for survivorship-facing or clinical decision support applications, where both completeness and correctness are critical.

\subsection{Impact of List Size Constraints}
We evaluated the effect of imposing a list size constraint by instructing models to generate 20--30 radiation side effects, without providing a predefined side-effect list. This prompting strategy is commonly used to encourage completeness, but may also incentivize fabrication when models are required to meet a target list length. Table 2 reports changes in performance relative to unconstrained free-form generation.

\begin{table*}[t]
\floatconts  {tab:numeric-delta}
\centering
  {\small\begin{tabular}{lccc}
  \toprule
  \bfseries Model & \bfseries $\Delta$Precision & \bfseries $\Delta$Recall & \bfseries $\Delta$F1 \\
  \midrule
  LLaMA-3.1-8B        & -0.09 (-0.12, -0.06) & -0.15 (-0.23, -0.08) & -0.13 (-0.16, -0.10) \\
  Gemma-2-9B          & -0.52 (-0.57, -0.47) &  0.03 (0.00, 0.06) & -0.10 (-0.13, -0.07) \\
  Mistral-7B          & -0.33 (-0.41, -0.27) & -0.08 (-0.14, -0.02) & -0.20 (-0.25, -0.15) \\
  Qwen-2.5-7B         & -0.27 (-0.33, -0.21) & -0.10 (-0.19, -0.01) & -0.22 (-0.28, -0.17) \\
  Phi-3-Medium-4K     & -0.48 (-0.59, -0.36) &  0.01 (-0.06, 0.07) & -0.20 (-0.26, -0.14) \\
  O4-Mini             & -0.20 (-0.25, -0.15) &  0.07 (0.03, 0.10) & -0.05 (-0.09, -0.02) \\
  GPT-5.3             & -0.21 (-0.24, -0.18) &  -0.02 (-0.05, -0.01) & -0.10 (-0.13, -0.07) \\ 
  \bottomrule
  \end{tabular}}
  {\caption{Effect of List Size Constraints (20–30 Side Effects) Relative to Unconstrained Free-Form Generation. Values reported as mean difference (95\% bootstrapped CI) across 21 patient profiles. Positive values indicate improvement; negative values indicate decline.}}
\end{table*}

Across all evaluated models, constraints on the number of side effects generated consistently reduced precision, with declines ranging from 0.09 to 0.52. In contrast, recall exhibited only small and inconsistent changes, with slight gains for some models and slight decreases for others. As a result, F1 scores generally decreased, reflecting a disproportionate loss in precision that was not offset by improved coverage of clinically relevant side effects.

These findings indicate that list size constraints do not reliably improve recall in free-form generation settings and instead tend to increase the inclusion of incorrect or marginally relevant side effects. In high-stakes clinical contexts such as oncology survivorship care, imposing a target list length alone is therefore insufficient to guide LLMs toward more reliable or clinically appropriate outputs.

\subsection{Grounded Selection from Clinician-Curated Lists}
To assess whether grounding improves reliability, we repeat experiments in which models are provided with a clinician-curated list of radiation-related side effects and asked to select those relevant to a given patient profile. We evaluate both unconstrained selection and selection under a list size constraint (20--30 side effects).

\begin{table*}[t]
\centering
\floatconts
  {tab:grounded}
  {\small\begin{tabular}{lccc}
  \toprule
  \bfseries Model & \bfseries Precision & \bfseries Recall & \bfseries F1 \\
  \midrule
  LLaMA-3.1-8B         & 0.87 (0.83, 0.90) & 0.54 (0.47, 0.63) & 0.65 (0.60, 0.70) \\
  Gemma-2-9B           & 0.90 (0.85, 0.95) & 0.37 (0.34, 0.39) & 0.52 (0.49, 0.55) \\
  Mistral-7B           & 0.86 (0.82, 0.90) & 0.49 (0.46, 0.54) & 0.62 (0.59, 0.65) \\
  Qwen-2.5-7B          & 0.63 (0.48, 0.78) & 0.70 (0.62, 0.78) & 0.54 (0.46, 0.68) \\
  Phi-3-Medium-4K      & 0.83 (0.74, 0.88) & 0.52 (0.48, 0.55) & 0.64 (0.58, 0.67) \\
  O4-Mini              & 0.92 (0.88, 0.95) & 0.87 (0.85, 0.89) & 0.89 (0.87, 0.91) \\
  GPT-5.3              & 0.90 (0.87, 0.93) & 0.87 (0.84, 0.89) & 0.88 (0.86, 0.90) \\
  \bottomrule
  \end{tabular}}
  {\caption{Performance of LLMs in Selecting Radiation Side Effects from a Clinician-Curated List. Values reported as mean (95\% bootstrapped CI) across 21 patient profiles.}}
\end{table*}

As shown in Table 3, grounded selection substantially improves precision and generally improves recall across all evaluated models relative to free-form generation. Performance gains are especially pronounced for precision, indicating that restricting the output space mitigates hallucination and inclusion of clinically irrelevant items.

\begin{table*}[t]
\centering
\floatconts
  {tab:grounded-lsc}
  {\small\begin{tabular}{lccc}
  \toprule
  \bfseries Model & \bfseries Precision & \bfseries Recall & \bfseries F1 \\
  \midrule
LLaMA-3.1-8B    & 0.81 (0.74, 0.87) & 0.85 (0.79, 0.90) & 0.81 (0.77, 0.85) \\
Gemma-2-9B      & 0.89 (0.84, 0.93) & 0.61 (0.57, 0.65) & 0.71 (0.68, 0.75) \\
Mistral-7B      & 0.85 (0.82, 0.88) & 0.91 (0.88, 0.94) & 0.88 (0.86, 0.89) \\
Qwen-2.5-7B     & 0.22 (0.14, 0.31) & 0.86 (0.80, 0.91) & 0.31 (0.24, 0.40) \\
Phi-3-Medium-4K & 0.72 (0.59, 0.82) & 0.59 (0.51, 0.67) & 0.64 (0.54, 0.72) \\
O4-Mini         & 0.89 (0.86, 0.93) & 0.94 (0.92, 0.96) & 0.91 (0.90, 0.93) \\
GPT-5.3         & 0.85 (0.84, 0.90) & 0.86 (0.83, 0.89) & 0.86 (0.84, 0.88) \\
  \bottomrule
  \end{tabular}}
  {\caption{Performance of LLMs in selecting 20--30 radiation side effects from a clinician-curated list. Values reported as mean (95\% bootstrapped CI) across 21 patient profiles.}}
\end{table*}

Table 4 shows that applying a list size constraint in the grounded setting consistently increases recall, but its effect on precision is model-dependent. While most models maintain high precision, Qwen-2.5-7B exhibits pronounced over-selection when required to meet a target list length. Unlike free-form generation, list size constraints generally do not degrade performance when models are restricted to selecting from an expert-curated vocabulary.

These results demonstrate that grounding LLM outputs in clinician-curated side-effect lists meaningfully improves reliability and robustness, and that list size constraints are substantially safer when applied in conjunction with grounded selection.

\subsection{Sensitivity to Documentation Specificity}
Across prompting regimes, we evaluate robustness to clinically realistic documentation changes by computing overlap ratios between side-effect lists generated from base and specified patient profiles. We focus on free-form (FF) and list-size–constrained free-form (20--30 side effects) prompts, because these settings reflect true model behavior under unconstrained or minimally constrained generation. Overlap measures how consistently a model preserves side effects when minor documentation changes are introduced, which would be artificially high if outputs were selected from a fixed clinician-curated list.

\begin{table}[t]
\floatconts
  {tab:overlap-freeform-numeric}
    {\small\begin{tabular}{lcc}
    \toprule
    \bfseries Model & \bfseries Overlap (FF) & \bfseries Overlap (LSC) \\
    \midrule
    LLaMA-3.1-8B & 0.21 (0.20, 0.23) & 0.21 (0.17, 0.25) \\
    Gemma-2-9B & 0.68 (0.61, 0.75) & 0.63 (0.56, 0.70) \\
    Mistral-7B & 0.44 (0.39, 0.50) & 0.43 (0.35, 0.53) \\
    Qwen-2.5-7B & 0.26 (0.24, 0.29) & 0.36 (0.30, 0.42) \\
    Phi-3-Medium-4K & 0.34 (0.28, 0.40) & 0.37 (0.32, 0.41) \\
    O4-Mini & 0.48 (0.45, 0.50) & 0.44 (0.42, 0.47) \\
    GPT-5.3 & 0.53 (0.50, 0.56) & 0.49 (0.46, 0.52) \\
    \bottomrule
    \end{tabular}}
    {\caption{Overlap ratios between base and specified patient profiles. FF = free-form; LSC = list-size–constrained free-form (20–30 items). Low values indicate high sensitivity to documentation specificity. Values reported as mean (95\% bootstrapped CI) across 21 patient profiles.}}
  
\end{table}

We observe that overlap ratios remain low across models in both FF and list size-constrained FF settings, demonstrating that minor documentation changes such as specifying radiation location can substantially alter generated side-effect lists. List size constraints produce small changes in overlap for some models, but they do not fully stabilize outputs. This highlights a deployment risk: LLM outputs may be inconsistent even for identical patients when input documentation varies slightly.

\subsection{Survivorship-Relevant Recall Analysis}

To assess survivorship-relevant performance disparities, we stratified recall by two dimensions: temporal onset (short-term vs long-term) and side-effect frequency (common, uncommon, rare, extremely rare). Recall was averaged across all prompting regimes to simplify presentation while preserving key trends.

\subsubsection{Recall by Temporal Onset}
Table 6 shows the average recall for short-term versus long-term side effects across models.

\begin{table}[t]
  \resizebox{\columnwidth}{!}{
  \begin{tabular}{lcc}
  \toprule
  \bfseries Model & \bfseries Short-term & \bfseries Long-term \\
  \midrule
  LLaMA-3.1-8B & 0.61 (0.56, 0.65) & 0.54 (0.50, 0.58) \\
  Gemma-2-9B & 0.44 (0.39, 0.49) & 0.29 (0.25, 0.32) \\
  Mistral-7B & 0.59 (0.55, 0.64) & 0.49 (0.44, 0.53) \\
  Qwen-2.5-7B & 0.56 (0.52, 0.60) & 0.50 (0.45, 0.54) \\
  Phi-3-Medium-4K & 0.48 (0.42, 0.53) & 0.40 (0.36, 0.43) \\
  O4-Mini & 0.70 (0.65, 0.74) & 0.82 (0.79, 0.84) \\
  GPT-5.3 & 0.69 (0.65, 0.73) & 0.79 (0.76, 0.82) \\
  \bottomrule
  \end{tabular}
  }
  \caption{Average recall of LLMs for short-term vs long-term side effects. Recall is averaged across frequency levels and prompting regimes. Values reported as mean (95\% bootstrapped CI).}\label{tab:recall-temporal}
\end{table}

Across most models, recall is generally higher for short-term effects compared to long-term effects, reflecting an under-recall of delayed toxicities such as breast or chest wall fibrosis, telangiectasia, lymphedema, cardiotoxicity, and secondary malignancies. Notably, O4-Mini and GPT-5.3 reverse this pattern, achieving higher recall for long-term effects than short-term effects, suggesting stronger coverage of delayed toxicity concepts.

\subsubsection{Recall by Side-Effect Frequency}
The following table presents recall stratified by frequency category (common, uncommon, rare, extremely rare), averaged over temporal onset and prompting regimes.

\begin{table*}[t]
\floatconts
  {tab:recall-frequency}
  {\resizebox{\textwidth}{!}{
    \small\begin{tabular}{lcccc}
    \toprule
    \bfseries Model & \bfseries Common & \bfseries Uncommon & \bfseries Rare & \bfseries Extremely Rare \\
    \midrule
    LLaMA-3.1-8B & 0.63 (0.58, 0.67) & 0.63 (0.56, 0.69) & 0.53 (0.47, 0.58) & 0.41 (0.31, 0.48) \\
    Gemma-2-9B & 0.60 (0.54, 0.65) & 0.32 (0.26, 0.38) & 0.21 (0.17, 0.24) & 0.20 (0.14, 0.25) \\
    Mistral-7B & 0.66 (0.63, 0.73) & 0.57 (0.50, 0.64) & 0.39 (0.34, 0.44)  & 0.44 (0.35, 0.52) \\
    Qwen-2.5-7B & 0.68 (0.61, 0.70) & 0.56 (0.46, 0.60) & 0.48 (0.44, 0.55) & 0.37 (0.30, 0.44) \\
    Phi-3-Medium-4K & 0.75 (0.68, 0.77) & 0.37 (0.34, 0.47) & 0.23 (0.20, 0.31) & 0.19 (0.11,  0.25) \\
    O4-Mini & 0.79 (0.75, 0.82) & 0.70 (0.64, 0.78) & 0.48 (0.41, 0.53) & 0.80 (0.75, 0.86) \\
    GPT-5.3 & 0.82 (0.79, 0.84) & 0.83 (0.78, 0.87) & 0.57 (0.51, 0.62) & 0.72 (0.64, 0.80) \\
    \bottomrule
    \end{tabular}
  }}
  {\caption{Average recall of LLMs by side-effect frequency. Recall is averaged across temporal onset and prompting regimes. Values reported as mean (95\% bootstrapped CI).}}
\end{table*}

Models generally recall common side effects more reliably than rare or extremely rare toxicities. Under-recall is particularly pronounced for uncommon and extremely rare effects, which frequently include long-term complications that are critical for survivorship care. More specifically, many models failed to retrieve clinically important rare toxicities such as pericarditis, despite these being present in the expert-curated vocabulary. These results highlight that LLMs tend to surface frequently discussed or acute side effects while under-representing delayed or infrequent toxicities, even under grounding or list size constraints.

Combining Tables 6 and 7, it is evident that LLMs are biased toward recalling short-term and common effects, with persistent gaps in long-term and rare side-effect coverage. O4-Mini and GPT-5.3 are exceptions, showing relatively strong performance across both temporal and frequency dimensions. This underscores the importance of expert curation or structured output constraints for generating clinically reliable survivorship information.

\section{Limitations}
This study has several limitations that should be considered when interpreting the results. First, our evaluation is conducted on 21 breast cancer patient profiles derived from publicly available synthetic and benchmark datasets. While this design enables controlled stress testing and isolates the effect of documentation specificity, the small sample size limits statistical generalization. 

Additionally, although profiles are formatted to resemble EHR-style documentation, they do not capture the full heterogeneity, ambiguity, or longitudinal complexity of real-world oncology records. Radiation specificity is introduced by appending radiation type or anatomical location directly to the term ``radiation'' while holding all other profile content fixed. While this isolates documentation effects, it does not reflect how radiation details are expressed in real clinical notes (e.g., narratives, abbreviations, or implicit cues).

Our study is also limited to breast cancer radiation therapy, which was chosen due to its large survivor population and well-characterized treatment-specific toxicity profiles. While the proposed framework is methodologically extensible and can in principle be applied to other cancer types and treatment modalities, we do not empirically evaluate these settings in this work. 

A further limitation is that the overlap ratio captures only output stability under perturbations and does not distinguish clinically appropriate variation from spurious inconsistency. However, unlike accuracy-based evaluation, the overlap ratio is directly measurable and does not rely on a single ground-truth definition of relevant side effects, which can be inherently context-dependent.


We also employ an LLM-based semantic matching procedure to align generated side effects with the clinician-curated reference, allowing for variation in wording while requiring equivalence in clinical meaning. Our LLM-based semantic matching procedure may misclassify borderline cases, though manual validation on 30 examples found 96\% agreement with human judgments.

Lastly, while the task of side-effect list generation is motivated by real clinical use cases such as informed consent drafting and survivorship education, we do not evaluate model performance within a deployed clinical workflow or measure downstream clinical outcomes. Our evaluation instead focuses on the reliability and consistency of generated content under controlled conditions, which we view as a necessary precursor to safe real-world integration.

\section{Conclusions}
In this work, we present a stress-testing framework to measure how accurately LLMs identify side effects of breast cancer radiation treatments. By systematically varying documentation specificity and prompting strategies while holding patient information fixed, we isolate failure modes that are likely to arise under real-world use rather than benchmark-style evaluation.

Across seven instruction-tuned LLMs, we observe consistent reliability limitations. Free-form generation exhibits a trade-off between precision and recall: some models, such as LLaMA-3.1-8B, produce broader lists with moderate recall but very low precision, indicating substantial hallucination; other models, such as Gemma-2-9B and O4-Mini, achieve high precision but at the cost of missing many clinically relevant side effects. Models also demonstrate pronounced sensitivity to small but clinically realistic changes in documentation, leading to unstable outputs for otherwise identical patient profiles. Additionally, LLMs systematically under-recall rare and long-term radiation side effects, even when common acute effects are captured reliably, raising concerns for survivorship-focused applications.

In contrast, constraining model outputs to clinician-curated side-effect lists substantially improves precision, recall, and robustness across all evaluated models. These findings suggest that grounding LLMs in expert-authored vocabularies is a necessary design choice for safer deployment in oncology information tasks, particularly those supporting survivorship care and patient education.

More broadly, our results highlight the importance of application-driven evaluation of medical LLMs. Evaluations that ignore documentation variability, prompting practices, and clinically salient distinctions may substantially overestimate real-world reliability. Factual accuracy and robustness under controlled perturbations represent a necessary first step toward clinical utility, and we therefore view such stress-testing frameworks as preceding more complex evaluations workflow integration and downstream clinical impact assessment. We hope this work motivates rigorous, deployment-aware stress testing of LLMs in high-stakes clinical domains before their integration into patient-facing tools or clinical decision-support systems, ultimately advancing safe and effective use of AI to improve survivorship care.

\section{Acknowledgements}
We thank the University of Virginia's Research Computing team for supporting the experiments in this work.

\bibliography{chil-sample}

@article{tonorezos2024prevalence,
  title={Prevalence of cancer survivors in the United States},
  author={Tonorezos, Emily and Devasia, Theresa and Mariotto, Angela B and Mollica, Michelle A and Gallicchio, Lisa and Green, Paige and Doose, Michelle and Brick, Rachelle and Streck, Brennan and Reed, Crystal and others},
  journal={JNCI: Journal of the National Cancer Institute},
  volume={116},
  number={11},
  pages={1784--1790},
  year={2024},
  publisher={Oxford University Press}
}

@article{mollica2025defining,
  title={Defining concepts in cancer survivorship},
  author={Mollica, Michelle A and Doose, Michelle and Reed, Crystal and Tonorezos, Emily},
  journal={Cancer},
  volume={131},
  number={16},
  pages={e70039},
  year={2025}
}

@article{wagle2025cancer,
  title={Cancer treatment and survivorship statistics, 2025},
  author={Wagle, Nikita Sandeep and Nogueira, Leticia and Devasia, Theresa P and Mariotto, Angela B and Yabroff, K Robin and Islami, Farhad and Jemal, Ahmedin and Alteri, Rick and Ganz, Patricia A and Siegel, Rebecca L},
  journal={CA: A cancer journal for clinicians},
  volume={75},
  number={4},
  pages={308--340},
  year={2025},
  publisher={Wiley Online Library}
}

@article{siegel2026cancer,
  title={Cancer statistics, 2026},
  author={Siegel, Rebecca L and Kratzer, Tyler B and Wagle, Nikita Sandeep and Sung, Hyuna and Jemal, Ahmedin},
  journal={Ca},
  volume={76},
  number={1},
  pages={e70043},
  year={2026}
}

@article{grunfeld2010interface,
  title={The interface between primary and oncology specialty care: treatment through survivorship},
  author={Grunfeld, Eva and Earle, Craig C},
  journal={Journal of the National Cancer Institute Monographs},
  volume={2010},
  number={40},
  pages={25--30},
  year={2010},
  publisher={Oxford University Press}
}

@article{ke2024decision,
  title={Decision aids for cancer survivors’ engagement with survivorship care services after primary treatment: a systematic review},
  author={Ke, Yu and Zhou, Hanzhang and Chan, Raymond Javan and Chan, Alexandre},
  journal={Journal of Cancer Survivorship},
  volume={18},
  number={2},
  pages={288--317},
  year={2024},
  publisher={Springer}
}

@inproceedings{alfano2022engaging,
  title={Engaging TEAM medicine in patient care: redefining cancer survivorship from diagnosis.},
  author={Alfano, Catherine M and Oeffinger, Kevin and Sanft, Tara and Tortorella, Brooke},
  booktitle={American Society of Clinical Oncology Educational book. American Society of Clinical Oncology. Annual Meeting},
  volume={42},
  pages={1--11},
  year={2022}
}

@article{nekhlyudov2017integrating,
  title={Integrating primary care providers in the care of cancer survivors: gaps in evidence and future opportunities},
  author={Nekhlyudov, Larissa and O'malley, Denalee M and Hudson, Shawna V},
  journal={The lancet oncology},
  volume={18},
  number={1},
  pages={e30--e38},
  year={2017},
  publisher={Elsevier}
}

@article{nathan2013family,
  title={Family physician preferences and knowledge gaps regarding the care of adolescent and young adult survivors of childhood cancer},
  author={Nathan, Paul Craig and Daugherty, Christopher Keller and Wroblewski, Kristen Elizabeth and Kigin, Mackenzie Louise and Stewart, Tom Vernon and Hlubocky, Fay Jarmila and Grunfeld, Eva and Del Giudice, Marie Elisabeth and Ward, Leigh-Anne Evelyn and Galliher, James Mahlon and others},
  journal={Journal of Cancer Survivorship},
  volume={7},
  number={3},
  pages={275--282},
  year={2013},
  publisher={Springer}
}

@article{bitterman2024promise,
  title={Promise and perils of large language models for cancer survivorship and supportive care},
  author={Bitterman, Danielle S and Downing, Andrea and Mau{\'e}s, Julia and Lustberg, Maryam},
  journal={Journal of Clinical Oncology},
  volume={42},
  number={14},
  pages={1607},
  year={2024}
}

@article{chen2025medical,
  title={Medical accuracy of artificial intelligence chatbots in oncology: a scoping review},
  author={Chen, David and Avison, Kate and Alnassar, Saif and Huang, Ryan S and Raman, Srinivas},
  journal={The Oncologist},
  volume={30},
  number={4},
  pages={oyaf038},
  year={2025},
  publisher={Oxford University Press US}
}

@misc{yoon2025navigating,
  title={Navigating artificial intelligence (AI) accuracy: A meta-analysis of hallucination incidence in large language model (LLM) responses to oncology questions.},
  author={Yoon, Sung Mi and Lyu, Jiyon and Djunadi, Trie Arni and Song, Junmin and Kim, Hye Sung and Min, Ronald Seungjune and Sakellakis, Minas and Chae, Young Kwang},
  year={2025},
  publisher={American Society of Clinical Oncology}
}

@article{singhal2023large,
  title={Large language models encode clinical knowledge},
  author={Singhal, Karan and Azizi, Shekoofeh and Tu, Tao and Mahdavi, S Sara and Wei, Jason and Chung, Hyung Won and Scales, Nathan and Tanwani, Ajay and Cole-Lewis, Heather and Pfohl, Stephen and others},
  journal={Nature},
  volume={620},
  number={7972},
  pages={172--180},
  year={2023},
  publisher={Nature Publishing Group UK London}
}

@article{sung2021global,
  title={Global cancer statistics 2020: GLOBOCAN estimates of incidence and mortality worldwide for 36 cancers in 185 countries},
  author={Sung, Hyuna and Ferlay, Jacques and Siegel, Rebecca L and Laversanne, Mathieu and Soerjomataram, Isabelle and Jemal, Ahmedin and Bray, Freddie},
  journal={CA: a cancer journal for clinicians},
  volume={71},
  number={3},
  pages={209--249},
  year={2021},
  publisher={Wiley Online Library}
}

@article{boyages2018evolution,
  title={Evolution of radiotherapy techniques in breast conservation treatment},
  author={Boyages, John and Baker, Lesley},
  journal={Gland Surgery},
  volume={7},
  number={6},
  pages={576},
  year={2018}
}

@article{palepu2025exploring,
  title={Exploring large language models for specialist-level oncology care},
  author={Palepu, Anil and Dhillon, Vikram and Niravath, Polly and Weng, Wei-Hung and Prasad, Preethi and Saab, Khaled and Tanno, Ryutaro and Cheng, Yong and Mai, Hanh and Burns, Ethan and others},
  journal={NEJM AI},
  volume={2},
  number={11},
  pages={AIcs2500025},
  year={2025},
  publisher={Massachusetts Medical Society}
}

@article{chen2023impact,
  title={The impact of using an AI chatbot to respond to patient messages},
  author={Chen, Shan and Guevara, Marco and Moningi, Shalini and Hoebers, Frank and Elhalawani, Hesham and Kann, Benjamin H and Chipidza, Fallon E and Leeman, Jonathan and Aerts, Hugo JWL and Miller, Timothy and others},
  journal={arXiv preprint arXiv:2310.17703},
  year={2023}
}

@article{ruiz2024leveraging,
  title={Leveraging large language models for precision monitoring of chemotherapy-induced toxicities: a pilot study with expert comparisons and future directions},
  author={Ruiz Sarrias, Oskitz and Mart{\'\i}nez del Prado, Mar{\'\i}a Purificaci{\'o}n and Sala Gonzalez, Mar{\'\i}a {\'A}ngeles and Azcuna Sagarduy, Josune and Casado Cuesta, Pablo and Figaredo Berjano, Covadonga and Galve-Calvo, Elena and L{\'o}pez de San Vicente Hern{\'a}ndez, Borja and L{\'o}pez-Santill{\'a}n, Mar{\'\i}a and Nu{\~n}o Escol{\'a}stico, Maitane and others},
  journal={Cancers},
  volume={16},
  number={16},
  pages={2830},
  year={2024},
  publisher={MDPI}
}

@article{chen2025large,
  title={Large language models in oncology: a review},
  author={Chen, David and Parsa, Rod and Swanson, Karl and Nunez, John-Jose and Critch, Andrew and Bitterman, Danielle S and Liu, Fei-Fei and Raman, Srinivas},
  journal={BMJ oncology},
  volume={4},
  number={1},
  pages={e000759},
  year={2025}
}

@article{mehan2025development,
  title={Development and evaluation of large-language models (LLMs) for oncology: A scoping review},
  author={Mehan, Namya and Desinghe, Teshan Dias and Saha, Ashirbani},
  journal={PLOS Digital Health},
  volume={4},
  number={8},
  pages={e0000980},
  year={2025},
  publisher={Public Library of Science San Francisco, CA USA}
}

@article{longwell2024performance,
  title={Performance of large language models on medical oncology examination questions},
  author={Longwell, Jack B and Hirsch, Ian and Binder, Fernando and Gonzalez Conchas, Galileo Arturo and Mau, Daniel and Jang, Raymond and Krishnan, Rahul G and Grant, Robert C},
  journal={JAMA Network Open},
  volume={7},
  number={6},
  pages={e2417641},
  year={2024}
}

@article{carl2024large,
  title={Large language model use in clinical oncology},
  author={Carl, Nicolas and Schramm, Franziska and Haggenm{\"u}ller, Sarah and Kather, Jakob Nikolas and Hetz, Martin J and Wies, Christoph and Michel, Maurice Stephan and Wessels, Frederik and Brinker, Titus J},
  journal={NPJ Precision Oncology},
  volume={8},
  number={1},
  pages={240},
  year={2024},
  publisher={Nature Publishing Group UK London}
}

@article{hao2025large,
  title={Large language model integrations in cancer decision-making: a systematic review and meta-analysis},
  author={Hao, Yuexing and Qiu, Zhiwen and Holmes, Jason and L{\"o}ckenhoff, Corinna E and Liu, Wei and Ghassemi, Marzyeh and Kalantari, Saleh},
  journal={NPJ Digital Medicine},
  volume={8},
  number={1},
  pages={450},
  year={2025},
  publisher={Nature Publishing Group UK London}
}

@article{kumar2025cross,
  title={A cross-sectional study of GPT-4--based plain language translation of clinical notes to improve patient comprehension of disease course and management},
  author={Kumar, Anivarya and Wang, Huanfei and Muir, Kelly W and Mishra, Vishala and Engelhard, Matthew},
  journal={Nejm Ai},
  volume={2},
  number={2},
  pages={AIoa2400402},
  year={2025},
  publisher={Massachusetts Medical Society}
}

@article{shi2025transforming,
  title={Transforming informed consent generation using large language models: mixed methods study},
  author={Shi, Qiming and Luzuriaga, Katherine and Allison, Jeroan J and Oztekin, Asil and Faro, Jamie M and Lee, Joy L and Hafer, Nathaniel and McManus, Margaret and Zai, Adrian H},
  journal={JMIR Medical Informatics},
  volume={13},
  number={1},
  pages={e68139},
  year={2025},
  publisher={JMIR Publications Inc., Toronto, Canada}
}

@article{ferber2025development,
  title={Development and validation of an autonomous artificial intelligence agent for clinical decision-making in oncology},
  author={Ferber, Dyke and El Nahhas, Omar SM and W{\"o}lflein, Georg and Wiest, Isabella C and Clusmann, Jan and Le{\ss}mann, Marie-Elisabeth and Foersch, Sebastian and Lammert, Jacqueline and Tschochohei, Maximilian and J{\"a}ger, Dirk and others},
  journal={Nature cancer},
  volume={6},
  number={8},
  pages={1337--1349},
  year={2025},
  publisher={Nature Publishing Group US New York}
}

@article{gegechkori2017long,
  title={Long-term and latent side effects of specific cancer types},
  author={Gegechkori, Nana and Haines, Lindsay and Lin, Jenny J},
  journal={Medical Clinics},
  volume={101},
  number={6},
  pages={1053--1073},
  year={2017},
  publisher={Elsevier}
}

@article{harrington2017late,
  title={The late medical effects of cancer treatments: a growing challenge for all medical professionals},
  author={Harrington, Jennifer and White, Jeff},
  journal={Clinical Medicine},
  volume={17},
  number={2},
  pages={137--139},
  year={2017},
  publisher={Elsevier}
}

@article{vos2024primary,
  title={Primary care physicians’ knowledge and confidence in providing cancer survivorship care: a systematic review},
  author={Vos, Julien AM and Wollersheim, Barbara M and Cooke, Adelaide and Ee, Carolyn and Chan, Raymond J and Nekhlyudov, Larissa},
  journal={Journal of Cancer Survivorship},
  volume={18},
  number={5},
  pages={1557--1573},
  year={2024},
  publisher={Springer}
}

@article{ross2022still,
  title={Still lost in transition? Perspectives of ongoing cancer survivorship care needs from comprehensive cancer control programs, survivors, and health care providers},
  author={Ross, Leslie W and Townsend, Julie S and Rohan, Elizabeth A},
  journal={International journal of environmental research and public health},
  volume={19},
  number={5},
  pages={3037},
  year={2022},
  publisher={MDPI}
}

@article{wang2021radiation,
  title={Radiation therapy-associated toxicity: Etiology, management, and prevention},
  author={Wang, Kyle and Tepper, Joel E},
  journal={CA: a cancer journal for clinicians},
  volume={71},
  number={5},
  pages={437--454},
  year={2021},
  publisher={Wiley Online Library}
}

@article{siau2021non,
  title={Non-oncologist physician knowledge of radiation therapy at an urban community hospital},
  author={Siau, Evan and Salazar, Hernando and Livergant, Jonathan and Klein, Jonathan},
  journal={Journal of Cancer Education},
  volume={36},
  number={1},
  pages={199--206},
  year={2021},
  publisher={Springer}
}

@article{kivistik2025perceptions,
  title={Perceptions, educational expectations and knowledge gaps of patients with non-metastatic breast cancer regarding radiotherapy: Integrative review},
  author={Kivistik, Siret and Mets{\"a}l{\"a}, Eija and Virtanen, Heli},
  journal={Technical Innovations \& Patient Support in Radiation Oncology},
  volume={34},
  pages={100312},
  year={2025},
  publisher={Elsevier}
}

@article{citrin2026effects,
  title={Effects of radiotherapy in normal tissue},
  author={Citrin, Deborah E and Timmerman, Robert D},
  journal={New England Journal of Medicine},
  volume={394},
  number={10},
  pages={996--1009},
  year={2026},
  publisher={Mass Medical Soc}
}

@article{jagsi2015differences,
  title={Differences in the acute toxic effects of breast radiotherapy by fractionation schedule: comparative analysis of physician-assessed and patient-reported outcomes in a large multicenter cohort},
  author={Jagsi, Reshma and Griffith, Kent A and Boike, Thomas P and Walker, Eleanor and Nurushev, Teamour and Grills, Inga S and Moran, Jean M and Feng, Mary and Hayman, James and Pierce, Lori J},
  journal={JAMA oncology},
  volume={1},
  number={7},
  pages={918--930},
  year={2015}
}

@article{shaitelman2015acute,
  title={Acute and short-term toxic effects of conventionally fractionated vs hypofractionated whole-breast irradiation: a randomized clinical trial},
  author={Shaitelman, Simona F and Schlembach, Pamela J and Arzu, Isidora and Ballo, Matthew and Bloom, Elizabeth S and Buchholz, Daniel and Chronowski, Gregory M and Dvorak, Tomas and Grade, Emily and Hoffman, Karen E and others},
  journal={JAMA oncology},
  volume={1},
  number={7},
  pages={931--941},
  year={2015}
}

\appendix

\section{Exact Prompts Used for LLM Evaluation}
\label{apd:prompts}

This appendix provides the exact prompts used for all prompting regimes
described in Section~3.4. In all cases, the placeholder
\texttt{\{\{PATIENT\_PROFILE\}\}} is replaced verbatim with the full
structured patient profile.

\subsection{Free-form Generation}

\begin{lstlisting}
I am going to give you a patient profile.

Your task is to list the short- and long-term side effects of the
patient's radiation treatment.

Output requirements:
- Output only a clean bullet-point list.
- Use one side effect per line.
- Do not include explanations, categories, commentary, or extra text.

{{PATIENT_PROFILE}}
\end{lstlisting}

\subsection{Free-form Generation with List Size Constraint (20-30)}

\begin{lstlisting}
I am going to give you a patient profile.

Your task is to list 20-30 short- and long-term side effects of the
patient's radiation treatment.

Output requirements:
- Output only a clean bullet-point list.
- Use one side effect per line.
- Do not include explanations, categories, commentary, or extra text.

{{PATIENT_PROFILE}}
\end{lstlisting}

\subsection{Selection from a Predefined Side-Effect List}

\begin{lstlisting}
I am going to give you a patient profile and a list of possible cancer
radiation side effects.

Your task is to identify which side effects from the list are relevant
to the patient's radiation treatment.

Output requirements:
- Output only a clean bullet-point list.
- Use one side effect per line.
- Do not include explanations, categories, commentary, or extra text.

{{PATIENT_PROFILE}}

Side effects:
brachial plexopathy
breast asymmetry
breast discomfort/sensitivity
breast or chest wall fibrosis
breast pain
breast shrinkage
breast swelling
cancer caused by radiation
cardiovascular disease
changes in color of the areola
compromised cosmetic result of reconstruction
decreased blood cell count
fatigue
hypothyroidism
lymphedema
mastitis
nausea
nipple pain
pericarditis
permanent hair loss in treated area
pneumonitis
poor wound healing
pulmonary fibrosis
radiation dermatitis
rib fracture
skin blistering or peeling
skin hyperpigmentation/changes in skin color
skin ulceration
sore throat
telangiectasia
temporary hair loss in the treated area
\end{lstlisting}

\subsection{Selection from a Predefined List with List size Constraint (20-30)}

\begin{lstlisting}
I am going to give you a patient profile and a list of possible cancer
radiation side effects.

Your task is to identify 20-30 side effects from the list that are
relevant to the patient's radiation treatment.

Output requirements:
- Output only a clean bullet-point list.
- Use one side effect per line.
- Do not include explanations, categories, commentary, or extra text.

{{PATIENT_PROFILE}}

Side effects:
brachial plexopathy
breast asymmetry
breast discomfort/sensitivity
breast or chest wall fibrosis
breast pain
breast shrinkage
breast swelling
cancer caused by radiation
cardiovascular disease
changes in color of the areola
compromised cosmetic result of reconstruction
decreased blood cell count
fatigue
hypothyroidism
lymphedema
mastitis
nausea
nipple pain
pericarditis
permanent hair loss in treated area
pneumonitis
poor wound healing
pulmonary fibrosis
radiation dermatitis
rib fracture
skin blistering or peeling
skin hyperpigmentation/changes in skin color
skin ulceration
sore throat
telangiectasia
temporary hair loss in the treated area
\end{lstlisting}

\end{document}